\documentclass[letterpaper]{article}
\usepackage{uai2018}
\usepackage[margin=1in]{geometry}

\usepackage{times}
\usepackage[noend]{algcompatible}
\usepackage{algorithm}
\usepackage{amsthm} % added by Noman for Definitions
\usepackage{amssymb}
\usepackage{mathtools}

\newtheorem{theorem}{Theorem}
\newtheorem{lemma}{Lemma}
\newtheorem*{lemma*}{Lemma}
\newtheorem{corollary}{Corollary}
\theoremstyle{definition}
\newtheorem{definition}{Definition}
\usepackage{caption,subcaption}

\DeclareMathOperator*{\argmax}{\text{arg}\max}

\newcommand{\MAX}{\texttt{MAX} }
\newcommand{\SUM}{\texttt{SUM} }
\newcommand{\eq}{equivalence }

\newcommand{\specialcell}[2][l]{%
  \begin{tabular}[#1]{@{}l@{}}#2\end{tabular}}
\usepackage{chngcntr}
\usepackage{graphicx}
\usepackage{bm}
\newcommand{\eat}[1]{}

\title{Lifted Marginal MAP Inference\thanks{\hspace{0.15cm}Paper accepted in UAI-18 (Sharma et al. 2018).}}

\author{Vishal Sharma\textsuperscript{1}, Noman Ahmed Sheikh\textsuperscript{2}, Happy Mittal\textsuperscript{1}, Vibhav Gogate\textsuperscript{3} \and Parag Singla\textsuperscript{1}\\
\textsuperscript{1}IIT Delhi, \{vishal.sharma, happy.mittal, parags\}@cse.iitd.ac.in\\
\textsuperscript{2}IIT Delhi, nomanahmedsheikh@outlook.com\\
\textsuperscript{3}UT Dallas, vgogate@hlt.utdallas.edu}

\begin{document}

\maketitle
\begin{abstract}
%Statistical Relational Learning (SRL) Models such as Markov logic combine the power of relational models (e.g., first-order logic) with probabilistic models (e.g., Markov networks) to represent underlying relational structure as well as handle uncertainty. 
Lifted inference reduces the complexity of inference in relational probabilistic models by identifying groups of constants (or atoms) which behave symmetric to each other. A number of techniques have
been proposed in the literature for lifting marginal as well MAP inference. We present the first application of lifting rules for marginal-MAP (MMAP), an important inference problem in
models having latent (random) variables. Our main contribution is two fold: (1) we define a new equivalence class of (logical) variables, called Single Occurrence for MAX (SOM), and show that solution lies at extreme
with respect to the SOM variables, i.e., predicate groundings differing only in the instantiation of the SOM variables take the same truth %(true/false) 
value (2) we define a sub-class {\em SOM-R} (SOM
Reduce) %of SOM 
and exploit properties of extreme assignments to show that MMAP inference can be performed by reducing the domain of SOM-R variables to a single constant. 
We refer to our lifting technique as the {\em SOM-R} rule for lifted MMAP.
%using the properties of extreme
%assignments, we show that MMAP problem can be solved over an equivalent theory in which the domain of SOM variables has been reduced to a single constant. We refer to our technique as the {\em
%we prove a general result showing that when the solution lies at extreme with respect a variable equivalence class, the MMAP assignment can be computed by reducing the domain of the variables in the equivalence class to a single constant.
%As a corollary, we state our {\em SOM} rule: the domain of SOM variables can be reduced to a single constant for MMAP inference. 
Combined with existing rules such as decomposer and binomial, this results in a powerful framework for lifted MMAP. Experiments on three benchmark domains show significant gains in both time and memory compared to
ground inference as well as lifted approaches not using SOM-R.
\end{abstract}

\vspace{-0.5cm}
\section{INTRODUCTION}
\vspace{-0.25cm}
Several real world applications such as those in NLP, vision and biology need to handle non-i.i.d. data as well as represent uncertainty. Relational Probabilistic models~\cite{getoor&taskar07} such as
Markov logic networks~\cite{domingos&lowd09} combine the power of relational representations with statistical models to achieve this objective. The na\"{i}ve approach to inference in these domains
grounds the relational network into a propositional one and then applies existing inference techniques. This can often result in sub-optimal performance for a large number of applications since inference is
performed oblivious to the underlying network structure.
%approach can be sub-optimal in practice because inference is performed oblivious to the underlying relational structure. 

Lifted inference~\cite{kimmig&al15} overcomes this shortcoming by collectively reasoning about groups of constants (atoms) which are identical to each other. Starting with the work of
Poole~\cite{poole03}, a number of lifting techniques which lift propositional  inference to the first-order level have been proposed in literature. For instance, for marginal inference, exact
algorithms such as variable elimination and AND/OR search and approximate algorithms such as belief propagation and MCMC sampling have been lifted to the first-order level (cf.
\cite{braz&al05,gogate&domingos11,broeck&al11,kersting&al09,singla&domingos08,niepert12,venugopal&gogate12}).  More recently, there has been increasing interest in lifting MAP inference (both exact
and approximate)~\cite{sarkhel&al14,mittal&al14,mladenov&al14}. Some recent work has looked at the problem of approximate lifting i.e., combining together those constants (atoms) which are similar but not necessarily identical~\cite{broeck&darwiche13,singla&al14,sarkhel&al15}.

%Another direction of work has explored techniques for performing approximate lifting, namely methods that infer over atoms that are similar but not necessarily identical~\cite{broeck&darwiche13,singla&al14,sarkhel&al15}. 
%Another way to characterize lifted inference algorithms is whether they tightly integrate the lifting with inference~\cite{singla&domingos08,venugopal&gogate12} or whether they describe the lifting behavior independent of the underlying inference algorithm~\cite{jha&al10,bui&al13,niepert12}.

Despite a large body of work on lifted inference, to the best of our knowledge, there is no work on lifted algorithms for solving marginal maximum-a-posteriori (MMAP) queries. MMAP inference is ubiquitous in real-world domains, especially those having latent (random) variables. 
%One of the several motivating examples is activity recognition in which actors interact with various objects while performing actions. The goal is to find MAP assignment to the activity while marginalizing over all other variables. Problems from vision, NLP and biology also have repeated structure and can be modeled using MLNs~\cite{maaten&al11}.
It is well known that in many real-world domains, the use of latent (random) variables significantly improves the prediction accuracy ~\cite{maaten&al11}. Moreover, the problem also shows up in the context of
SRL domains in tasks such as plan and activity recognition~\cite{singla&mooney11}. Therefore, efficient lifted methods for solving the MMAP problem are quite desirable. 

MMAP inference is much harder than marginal (sum) and MAP (max) inference because sum and max operators do not commute. In particular, latent (random) variables need to be marginalized out before MAP assignment can be computed over the query (random) variables and as a result MMAP is NP-hard even on tree graphical models \cite{park02}. 
%Even on tree graphical models, where both marginal and MAP inference are tractable, MMAP can be NP-hard~\cite{liu&ihler13}. 
Popular approaches for solving MMAP include variational algorithms~\cite{liu&ihler13}, AND/OR search~\cite{marinescu&al14} and parity solvers~\cite{xue&al16}.
%cite Radu's thesis?

In this paper, we propose the first ever lifting algorithm for MMAP by extending the class of lifting rules~\cite{jha&al10,gogate&domingos11,mittal&al14}. As our first contribution, we define a new
equivalence class of (logical) variables called {\em Single Occurrence for \MAX (SOM)}.
%with properties: (a) no two variables in the equivalence class appear in the same formula (2) at least
%one variable in the class appears in a \MAX predicate.
%extending the Single Occurrence (SO) variables of Mittal et al.~\shortcite{mittal&al14}, we propose a new equivalence class of variables called {\em Single Occurrence for MAX (SOM)}. Intuitively, in addition to the
%SO condition of Mittal et al., SOM also requires at least one variable in the class to appear
%in a \texttt{MAX} predicate. 
We show that the MMAP solution
lies at extreme with respect to the SOM variables, i.e., predicate groundings which differ only in the instantiation of the SOM variables take the same truth (true/false) value in the MMAP assignment.
The proof is fairly involved due to the presence of both \texttt{MAX} and \texttt{SUM} operations in MMAP,
%(which do not commute with each other), 
and involves a series of problem transformations followed by exploiting the convexity of the resulting function.
%to arrive at the desired result. 
%followed by exploiting the convexity of the resulting function. 

As our second contribution, we define a sub-class of SOM, referred to as {\em SOM-R} (SOM Reduce). Using the properties of extreme assignments, we show that the MMAP solution can be computed by
reducing the domain of SOM-R variables to a single constant.
%we exploit the properties of extreme assignments to show that the MMAP solution can be computed over an equivalent theory in which the domain of SOM variables has been
%reduced to a single constant. 
We refer to this as {\em SOM-R} rule for lifted MMAP.
%we prove that whenever solution lies at extreme with respect to a variable equivalence class, the MMAP assignment can be computed by solving a simpler theory where the
%domain of the equivalence class variables has been reduced to a single constant. As a corollary, we deduce that MMAP problem can be equivalently solved by reducing the domain of SOM variables to a single constant. 
%We refer to this as our {\em SOM} rule for MMAP. 
SOM-R rule is often applicable when none of the other rules are, and can result in significant savings since inference complexity is exponential in the domain size in the worst case. 

Finally, we show how to combine SOM-R rule along with other lifting rules e.g., binomial and decomposer, resulting in a powerful algorithmic framework for lifted MMAP inference. Our experiments on three
different benchmark domains clearly demonstrate that our lifting technique can result in orders of magnitude savings in both time and memory compared to ground inference as well as vanilla lifting
(not using the SOM-R rule).

\vspace{-0.25cm}
\section{BACKGROUND}\label{sec:background}
\vspace{-0.25cm}
{\bf First-Order Logic:} The language of first-order logic~\cite{russell&norvig10} consists of {\em constant}, 
{\em variable}, {\em predicate}, and {\em function} symbols. A {\em term} 
is a variable, constant or is obtained by application of a function to a tuple of terms.
\nocite{sharma&al18}
Variables in first-order logic are often referred to as {\em logical variables}. We will simply refer to them as variables, henceforth. A {\em predicate} defines a relation over the set of its arguments. 
An {\em atom} is obtained by applying a predicate symbol to the corresponding arguments. A {\em ground atom} is an atom having no variables in it.
% A {\em ground atom} is a predicate having no variables in its arguments.
Formulas are obtained by combining predicates using a set operators: $\wedge$ (and), 
$\vee$ (or) and $\neg$ (not). Variables in a formula can be universally or existentially 
quantified using the operators $\forall$ and $\exists$, respectively. A first-order theory (knowledge 
base) is a set of formulas. We will restrict our attention to function free finite
first-order logic with Herbrand interpretation~\cite{russell&norvig10} and universally quantified variables.
%, as assumed by most earlier work in this domain~\cite{domingos&lowd09}. 
%We will also restrict our attention to the case of universally quantified variables. 
In the process of (partially) grounding a theory, we replace all (some) of the universally 
quantified variables with the possible constants in the domain. In the following, we will use capital
letters (e.g., $X$, $Y$ etc.) to denote logical variables and small case letters to denote constants. We will use $\Delta_X=\{x_1,x_2,\cdots,x_m\}$ denotes the domain of variable $X$.

{\bf Markov Logic:}
A Markov logic network~\cite{domingos&lowd09} (MLN) $M$ is defined as a set of pairs $\{f_i,w_i\}_{i=1}^n$ 
where $f_i$ is a formula in first-order logic and $w_i$ is the weight of $f_i$.
We will use $F(M)$ to denote the set of all the formulas in MLN.
%Let $\mathcal{S}$ denote the set of all the predicates in $M$. 
Let $\mathcal{X}$ denote the set of all the logical variables appearing in MLN. An MLN can be seen as a template for constructing ground Markov networks. Given the domain $\Delta_X$ for every variable
$X \in \mathcal{X}$, the ground network constructed by MLN has a node for every ground atom and a feature for every ground formula. Let $\mathcal{T}$ denote the set of all the predicates appearing in
$M$. We will use $\mathcal{T}_g$ to denote all the ground atoms corresponding to the set $\mathcal{T}$ and $t$ to denote an assignment, i.e. a vector of true/false values, to $\mathcal{T}_g$. The distribution specified by an MLN is given
as:
\begin{align}
P(\mathcal{T}_g=t)= \frac{1}{Z} e^{\sum_{i=1}^{n} \sum_{j=1}^{m_i} w_i f_{ij}(t)}
\end{align}
where $m_i$ denotes the number of groundings of the $i^{th}$ formula. 
%where $i$ varies over the first-order logic formulas in the MLN and $j$ varies over the groundings of
%the $i^{th}$ formula. 
$f_{ij}$ represents the feature corresponding to the $j^{th}$ grounding of the $i^{th}$ formula. The feature is on if the corresponding formula is satisfied under the assignment $t$ off otherwise. $Z$ is the normalization constant. Equivalently, in the potential function representation, the distribution can be written as:
\begin{align}
\label{eqn:mln}
P(t)=\frac{1}{Z} \prod_{i=1}^{n} \prod_{j=1}^{m_i} \phi_{ij}(t)
\end{align}
where there is a potential $\phi_{ij}$ for each $f_{ij}$ such that $\phi_{ij}(t)=e^{w_i f_{ij}(t)}$.

{\bf Marginal MAP (MMAP):}
Let the set of all predicates $\mathcal{T}$ be divided into two disjoint subsets $\mathcal{Q}$ and $\mathcal{S}$, referred to as \texttt{MAX} and \texttt{SUM} predicates, respectively. Let $q$ (resp. $s$) denote an assignment to all the groundings of the predicates in $\mathcal{Q}$ (resp. $\mathcal{S}$). Note that $\mathcal{T}=\mathcal{Q} \cup 
\mathcal{S}$, and given assignment $t$ to $\mathcal{T}$, $t=q \cup s$. Then, the {\em marginal-MAP (MMAP)} problem for MLNs can be defined as:
\begin{align}
\label{eqn:mmap}
	&\argmax_{q} \sum_{s}\prod_{i=1}^{n} \prod_{j=1}^{m_i} \phi_{ij}(q, s) = \argmax_{q} W_M(q) \\
	&\text{where, } W_M(q)=\sum_{s}\prod_{i=1}^{n} \prod_{j=1}^{m_i} \phi_{ij}(q, s) \nonumber	
\end{align}
$W_M(q)$ is referred to as the MMAP objective function for the MLN $M$, and its solution $q^{*}=\argmax_q W_M(q)$ 
is referred as the MMAP solution.
%after summing out marginal
%variables (unnormalized) probability obtained at the assignment $q^g$ to the MAP predicates.
Note that we can get rid of $Z$ in equation~\ref{eqn:mmap}, since we are only interested in finding
the maximizing assignment and $Z$ is a constant. 

{\bf Preliminaries:}
%Consider an MLN $M$ given as $\{f_i,w_i\}_{i=1}^{n}$. 
%We will use capital letters e.g., $X,Y,Z$ to
%denote the set of logical variables in a formula. We will use $\Delta_X=\{x_1,x_2,\cdots,x_m\}$
%to denote the domain of variable $X$. We will use small case letters to denote constants.
%Given a variable $X$, we will use $\Delta_X$ to 
%denote its domain. We will use the small case letters e.g., $x_1,x_2,\cdots,x_m$ to denote the 
%individual elements in $\Delta_X$. 
%Given an MLN $M$ given as $\{f_i,w_i\}_{i=1}^{n}$, 
We will assume that our MLN is in Normal Form~\cite{mittal&al14} i.e., (a) no constants appear 
in any of the formulae (b) if $X$ and $Y$ appear at the same predicate position in one or more formulae, 
then $\Delta_X=\Delta_Y$. Any MLN can be converted into normal form by a series of mechanical operations. 
We will also assume that formulas are standardized apart i.e., we rename the variables such that the sets 
of variables appearing in two different formulae are disjoint with each other. We define an equivalence 
relation $\sim$ over the set of variables such that $X \sim Y$ if (a) $X$ and $Y$ appear at the same 
predicate position OR (b) $\exists Z$ such that $X \sim Z$ and $Y \sim Z$.
% To take an example, consider an MLN $M_1$ with two formulae (1) $w_1$ : $Q(X,Y) \wedge R(X,Z) \Rightarrow T(Y)$ (2) $w_2$: $T(U)$. 
% Then, $\{X\}$, $\{Y,U\}$ and $\{Z\}$ are the equivalence classes in this MLN. We will use $M_1$ as a working example for this paper.
We will use $\tilde{X}$ to denote the equivalence class corresponding to variable $X$. Variables in the same equivalence class must have the same domain due to the normal form assumption. We will use $\Delta_{\tilde{X}}$ to refer to the domain of the variables belonging to $\tilde{X}$.

Finally, though our exposition in this work is in terms of MLNs, our ideas can easily be generalized to other representations such as weighted  parfactors~\cite{braz&al05} and probabilistic knowledge bases~\cite{gogate&domingos11}.

\vspace{-0.25cm}
\section{SINGLE OCCURRENCE FOR MMAP}
\vspace{-0.25cm}
\subsection{Motivation}
\vspace{-0.15cm}
In this work, we are interested in lifting the marginal-MAP (MMAP) problem. Since MMAP is a problem harder than both marginal and MAP inference, a natural question to examine would be if existing
lifting techniques for MAP and marginal inference can be extended to the case of MMAP. Or further still, if additional rules can be discovered for lifting the MMAP problem. Whereas many of the
existing rules such as decomposer and binomial~\footnote{applicable when the binomial predicate belongs to \MAX}~\cite{jha&al10,mittal&al15}
%and the domain recursion~\cite{broeck&al11} 
extend in a straightforward manner for MMAP, unfortunately the SO rule~\cite{mittal&al14}, which is a powerful rule for MAP inference, is not directly applicable.

In response, we propose a new rule, referred to as Single Occurrence for MAX Reduce (SOM-R), which is applicable for MMAP inference. We first define a variable equivalence class, referred to as SOM, 
which requires that (1) no two variables in the class appear in the same formula (2) at least one of the variables in the class appears in a \MAX predicate. We further define a sub-class of
SOM, referred to as SOM-R, which imposes a third condition (3) either all the \SUM predicates in the theory contain a SOM variable or none of them does. Our SOM-R rule states that domain of SOM-R variables can be reduced to a single constant for MMAP inference. Consider the following example MLN, henceforth referred to as $M_1$:
\begin{align*}
	&w_1 : \mathit{Frnds}(X,Y) \wedge \mathit{Parent}(Z,X) \Rightarrow \mathit{Knows}(Z,Y) \\
	&w_2 : \mathit{Knows}(U,V) \\
    &\mathit{SUM:Parent \quad MAX: Frnds, Knows}
\end{align*}
The equivalence classes in this example are given by $\{X\}$, $\{Y,V\}$ and$\{Z, U\}$. It is easy to see that each of these equivalence classes satisfy the three conditions above and hence, SOM-R
rule can be applied over them. This makes the MMAP inference problem independent of the size of the domain and hence, it can be solved in $O(1)$ time. Ground inference has to deal with $O(m^2)$ number of
ground atoms resulting in $O(\exp(cm^{2}))$ complexity in the worst case~\footnote{Inference
complexity is exponential in the number of ground atoms. Here, we assume $|\Delta_X|=|\Delta_Y|=|\Delta_Z|=m$}, where $c$ is a constant. Further, in the absence of the SOM-R rule, none of the existing lifting rules apply and one has to resort to partial grounding again resulting in worst case exponential complexity.

We note that conditions for identifying SOM and SOM-R specifically make use of the structure of the MMAP problem. Whereas condition 1 is same as Mittal et al.'s SO condition, condition 2
requires the variables in the SOM class to belong to a \MAX predicate. Condition 3 (for SOM-R) further refines the SOM conditions so that domain reduction can be applied. 

We prove the correctness of our result in two phases. First, we show that SOM equivalence class implies that MMAP solution lies at extreme, meaning that predicate groundings differing only in the
instantiation of the SOM class take the same truth value. Second, for the sub-class SOM-R, we further show that domain can be reduced to a single constant for MMAP. Here, we rely on the properties of
extreme assignments.

%Non-commutative nature of \MAX and \SUM operators makes our proof of correctness fairly involved. 
Our proof strategy makes use of a series of problem transformations followed by using the convexity of the resulting function. These algebraic manipulations are essential to prove the correctness of
our result, and are some of the important contributions of our paper. Next, we describe each step in detail. The proofs of theorems (and lemmas) marked with ($*$) are in the supplement.
\vspace{-0.15cm}
\subsection{SOM implies Extreme Solution}
\label{sec:extreme}
\vspace{-0.15cm}
We introduce some important definitions. We will assume that we are given an MLN $M$. Further, we are interested in solving an MMAP problem over $M$ where the set of \MAX predicates is given by $\mathcal{Q}$.
\begin{definition}(Single Occurrence for MAX)
%Let $M$ be an MLN. Consider the MMAP problem over $M$ with set of \MAX predicates given as $\mathcal{Q}$. 
We say that a variable equivalence class $\tilde{X}$ is Single Occurrence for \MAX (SOM) if (a)
$\forall i, f_i \in F(M)$, there is at most one variable from the set $\tilde{X}$ occurring in $f_i$
(b) there exists a variable $X \in \tilde{X}$ and a predicate $P \in \mathcal{Q}$, such that $X$ appears in $P$.
\end{definition}
%In our example MLN $M_1$, $\{X\}$ is not SOM since $X$ does not appear in a \MAX predicate. 
Next, we define the notion of an extreme assignment.
\begin{definition}(Extreme Assignment)
	%Let $M$ be an MLN and 
	Let $\tilde{X}$ be a variable equivalence class. An assignment $q$ to \MAX predicates $\mathcal{Q}$ lies at extreme (with respect to $\tilde{X}$), if $\forall P \in \mathcal{Q}$, all the
	groundings of $P$ with the same instantiation to variables $\mathcal{X}-\tilde{X}$, take the same value in $q$. 
\end{definition}
In $M_1$, an extreme assignment with respect to variable equivalence class $\{Y,V\}$ will assign the same truth value to the ground atoms $\mathit{Knows(z,y_1)}$ and $\mathit{Knows(z,y_2)}$, $\forall
z \in \Delta_Z$ and $\forall y_1,y_2 \in \Delta_Y$. We next define the notion of an MLN variablized with respect to a variable equivalence class.
\begin{definition}\label{defn:variablize} (Variablized MLN)
%Let $M$ be an MLN. Let $\mathcal{X}$ denote the entire set of variables appearing in $M$ and let $\tilde{X}$ 
%be an equivalence class.
Let $\tilde{X}$ be an equivalence class. Let $M_{\tilde{X}}$ be the MLN obtained by instantiating (grounding) the variables in the set $\mathcal{X}-\tilde{X}$. We say that $M_{\tilde{X}}$ is variablized (only) with respect to 
the set $\tilde{X}$.
%i.e., all the variables in $M$ except those in the class $\tilde{X}$. 
%Then, we say that $M_{\tilde{X}}$ is partially ground % given the equivalence class $\tilde{X}$.
%with respect to the set $\mathcal{X}-\tilde{X}$. Alternately, 
\end{definition}
For instance in $M_1$, variablizing with respect to the equivalence class $\{Y,V\}$ results
in MLN with formulas similar to:
\begin{align*}
	&w_1 : \mathit{Frnds}(x,Y) \wedge \mathit{Parent}(z,x) \Rightarrow \mathit{Knows}(z,Y) \\
	&w_2 : \mathit{Knows}(u,V) 
\end{align*}
where $x,z$ and $u$ are constants belonging to respective domains. $\mathit{Frnds}(x,Y)$, ${\mathit{Knows}(z,Y)}$ and ${\mathit{Knows}(u,V)}$ can be treated as unary predicates over the equivalence class
$\{Y,V\}$ since $x,z$ and $u$ are constants. Similarly, $\mathit{Parent}(z,x)$ can be treated as a propositional predicate.

It is important to note that, $M_{\tilde{X}}$ represents the same distribution as $M$. Further, $M_{\tilde{X}}$
can be converted back into normal form by introducing a new predicate for every combination of constants 
appearing in a predicate. We now define one of the main theorems of this paper.
\begin{theorem}\label{thm:extreme}
Let $M$ be an MLN and let $\tilde{X}$ be a SOM equivalence class. Then, an MMAP solution for $M$ lies at extreme with respect to $\tilde{X}$.
\end{theorem}
We will prove the above theorem by defining a series of problem transformations. In the following, we will work with MLN $M$ and $\tilde{X}$ as a SOM variable equivalence class. We will use
$\mathcal{Q}$ and $\mathcal{S}$ to denote set of \MAX and \SUM predicates, respectively. $q$ and $s$ will denote the assignments to respective predicate groundings (see Background (section~\ref{sec:background})).
%In order to prove this theorem, we next define a series of problem transformation to prove the extrema property.

% \subsubsection{Problem Transformation 1 : \normalfont{(Variablizing $M$)}}
\vspace{-0.15cm}
\subsubsection{Problem Transformation (PT) 1}
\vspace{-0.15cm}
{\bf Objective PT1:} Convert MMAP objective into a form which only has unary and propositional predicates.
%In this transformation, we would like our resulting MLN to have only unary
%and propositional predicates. We can achieve this by variablizing our MLN w.r.t. any one of the SO classes (see Lemma ~\ref{lem:var_MLN} below).
\begin{lemma}
\label{lem:pt1}
Let $M_{\tilde{X}}$ denote the MLN variablized with respect to SOM equivalence class $\tilde{X}$. Then, $M_{\tilde{X}}$ contains only unary and propositional predicates. Further, the MMAP objective can be written as:
\begin{align*}
 \argmax_{q} W_M(q) = \argmax_{q} W_{M_{\tilde{X}}}(q)
\end{align*}
\end{lemma}
% \vspace{-0.5cm}
% The proof that $M_{\tilde{X}}$ only has unary and propositional predicates follows immediately from the definition of $M_{\tilde{X}}$ (defn. \ref{defn:variablize}) and the fact that $\tilde{X}$ is SOM.
The proof that $M_{\tilde{X}}$ only has unary and propositional predicates follows immediately from the definition of $M_{\tilde{X}}$ (defn. \ref{defn:variablize}) and the fact that $\tilde{X}$ is SOM.
Further, since $M$ and $M_{\tilde{X}}$ define the same distribution, we have the equivalence of the MMAP objectives. Since, $M_{\tilde{X}}$ only has unary and propositional predicates, we will split
the assignment $q$ to groundings of $\mathcal{Q}$ into $(q_u,q_p)$ where $q_u$ and $q_p$ denote the assignments to groundings of unary and propositional predicates, respectively. 
Similarly, for assignment $s$ to groundings of $\mathcal{S}$, we split $s$ as $(s_u,s_p)$.
\vspace{-0.15cm}
\subsubsection{Problem Transformation 2}
\vspace{-0.15cm}
{\bf Objective PT2:} In the MMAP objective, get rid of propositional \MAX predicates.
\begin{lemma}
\label{lem:pt2}
\hspace{-0.05in}{\bf *} Consider the MMAP problem over $M_{\tilde{X}}$. Let $q_{p}$ be some assignment to propositional \MAX predicates. Let $M^{\prime}_{ \tilde{X}}$ be an MLN obtained by
substituting the truth value in $q_{p}$ for propositional predicates. Then, if $M^{\prime}_{ \tilde{X}}$ has a solution at extreme for all possible assignments of the form $q_{p}$ then,
$M_{\tilde{X}}$ also has a solution at extreme. 
\end{lemma}
Therefore, in order to prove the extrema property for $M_{\tilde{X}}$, it is sufficient to prove
it for a generic MLN $M'_{\tilde{X}}$, i.e., without making any assumptions on the form of $q_{p}$. 

For ease of notation, we will drop the prime in $M'_{\tilde{X}}$ and simply refer to it as $M_{\tilde{X}}$.
%Similarly, we will use $q$ instead of using $q_u$. 
Therefore, we need to show that the solution to the following problem lies at extreme:
% \setcounter{equation}{4}
% $$\argmax_{q_u} W_{M_{\tilde{X}}}(q) $$
\begingroup
\begin{align*}
	\argmax_{q_u} W_{M_{\tilde{X}}}(q_u) 
\end{align*}
\endgroup
% \begin{align*}
% 	\argmax_{q_u} W_{M_{\tilde{X}}}(q) 
% \end{align*}\vspace{-0.1cm}
where the propositional \MAX predicates have been gotten rid of in $M_{\tilde{X}}$.
%Similarly, we will simply refer to $q_u$ simply as $q$ for notational convenience.
% 
% Problem Transformations
% 
\subsubsection{Problem Transformation 3}
\vspace{-0.15cm}
\label{sec:pt3}
{\bf Objective PT3:} In the MMAP objective, get rid of unary \SUM predicates using inversion elimination~\cite{braz&al05}.\\
%In this transformation, we will reformulate $W_{M'_{\tilde{X}}}(q_{u})$ to get rid of unary \SUM predicates by applying inversion elimination~\cite{braz&al06} leading to further simplification of the theory.
First, we note that the MMAP objective:
%\begin{equation}
% \setcounter{equation}{4}
\begin{align*}
& W_{M_{\tilde{X}}}(q_{u})  =  \sum_{s_{p}, s_{u}} \prod_{i=1}^{n} \prod_{j=1}^{m_i} \phi_{ij}(q_{u}, s_{p}, s_{u}) \nonumber\\
%\end{equation}
&  \text{can be equivalently written as:}  \nonumber \\
%\begin{equation}
& W_{M_{\tilde{X}}}(q_{u}) = \sum_{s_{p}, s_{u}} \prod_{i=1}^{n} \prod_{j=1}^{m} \phi'_{ij}(q_{u}, s_{p}, s_{u}) 
\end{align*}
%\end{equation}
where $m=|\Delta_{\tilde{X}}|$. $\phi'_{ij}(q_{u}, s_{p}, s_{u})=\phi_{ij}(q_{u}, s_{p}, s_{u})$ if $f_i$ contains a variable from 
$\tilde{X}$, else $\phi'_{ij}(q_{u}, s_{p}, s_{u})={\phi_{ij}(q_{u}, s_{p}, s_{u})}^{\frac{1}{m}}$
otherwise. It is easy to see this equivalence since the only variables in the theory are from the class
$\tilde{X}$. When $f_i$ contains a variable from $\tilde{X}$, it has exactly $m_i=m$ groundings since 
$\tilde{X}$ is SOM. On the other hand, if $f_i$ does not contain a variable from $\tilde{X}$, it only contains propositional predicates. Then we raise it to power $\frac{1}{m}$, and then multiply $m$
times in the latter expression to get an equivalent form.
% Next, we state an important lemma to get rid of unary \SUM predicates.

Next, we use inversion elimination~\cite{braz&al05} to get rid of unary \SUM predicates.
%which gets rid of unary \SUM predicates using inversion elimination~\cite{braz&al05}.
% The function $W_{M'_{\tilde{X}}}(q^{1g})$ can be written as function of $q^{1g},r^{0g}$ variables.
\begin{lemma}\label{lem:pt3}
\label{lem:inversion_elim}
MMAP problem over $M_{\tilde{X}}$ can be written as:
\begin{align*}
\argmax_{q_u} W_{M_{\tilde{X}}}(q_{u}) = \argmax_{q_u} \sum_{s_{p}}\prod_{j=1}^{m} \Theta_{j}(q_{u}, s_{p})
\end{align*}
where $\Theta_{j}$ is a function of unary \MAX and propositional \SUM predicates groundings $q_u$ and $s_p$, respectively.
	% The function $W_{M'_{\tilde{X}}}(q^{1g})$ can be written as function of $q^{1g},r^{0g}$ variables.
\end{lemma}
% \begin{proof}\renewcommand{\qedsymbol}{}
\textit{Proof.} We can write the MMAP objective $W_{M_{\tilde{X}}}(q_{u})$ as:
\begin{align}
&= \sum_{s_{p}, s_{u}} \prod_{i=1}^{n} \prod_{j=1}^{m} \phi'_{ij}(q_{u},s_{p}, s_{u})\nonumber\\
&= \sum_{s_{p}, s_{u}} \prod_{j=1}^{m} \prod_{i=1}^{n} \phi'_{ij}(q_{u},s_{p}, s_{u})\nonumber\\
 % &= \sum_{s_{p}, s_{u}} \prod_{j=1}^{m} \prod_{i=1}^{n} \phi'_{ij}(q_{u},s_{p}, s_{u})\nonumber\\
&=\sum_{s_{p}, s_{u}} \prod_{j=1}^{m} \Phi_{j}(q_{u},s_{p}, s_{u})\nonumber\\
&=\sum_{s_{p}}\sum_{s_{u_1}, s_{u_2},\hdots,s_{u_m}} \prod_{j=1}^{m} \Phi_{j}(q_{u},s_{p}, s_{u_j}) \nonumber\\
\end{align}
\begin{align}
& \text{(apply inversion elimination)} \nonumber \\
& =\sum_{s_{p}} \prod_{j=1}^{m} \sum_{s_{u_j}} \Phi_{j}(q_{u},s_{p}, s_{u_j})\nonumber\\
% & =\sum_{s_{p}} \prod_{j=1}^{m} \sum_{s_{u_j}} \Phi_{j}(q_{u},s_{p}, s_{u_j})&&\text{(inversion elimination)}\nonumber\\
&=\sum_{s_{p}}\prod_{j=1}^{m} \Theta_{j}(q_{u},s_{p})  \nonumber
\end{align}
% \end{proof}
{\bf Proof Explanation:} 
Second equality is obtained by interchanging the two products. Third equality is obtained by defining $\prod_i \phi'_{ij}(q_{u}, s_{p}, s_{u})=\Phi_j(q_{u}, s_{p}, s_{u})$. In fourth equality, we have made explicit the
dependence of $\Phi_j$ on $s_{u_j}$ i.e. the groundings corresponding to the $j^{th}$ constant. \\ 
{\bf Inversion Elimination}~\cite{braz&al06}{\bf :} Since $\Phi_j$ only depends on $s_{u_j}$ (among $s_{u}$) groundings, we can use inversion elimination to invert the sum over $s_{u_j}$ and product over
$j$ in the fifth equality. \\
{\bf Final Expression:} We define $\Theta_{j}(q_{u}, s_{p})=\sum_{s_{u_j}} \Phi_{j}(q_{u}, s_{p}, s_{u})$. \\
Note that, at this point, we have only propositional \SUM and unary \MAX predicates in the transformed MMAP objective.
\vspace{-0.15cm}
\subsubsection{Problem Transformation 4}
\vspace{-0.15cm}
\label{sec:pt4}
%{\bf Objective PT4:} Transform the MMAP objective in a form where convexity can be exploited \\
{\bf Objective PT4:} Exploit symmetry of the potential functions in the MMAP objective. \\
We rename $q_u$ to $q$ and $s_p$ to $s$ for ease of notation in Lemma~\ref{lem:pt3}. The MMAP objective can be written as:
\begin{align}
W_{M_{\tilde{X}}}(q) = \sum_{s}\prod_{j=1}^{m} \Theta_{j}(q_j, s) \label{eqn:pt4}
\end{align}
Here, $q=(q_1,q_2,\hdots,q_m)$ and $q_j$ represents the assignment to the unary $\MAX$ predicate groundings corresponding to constant $j$. In the expression above, we have made explicit the dependence of $\Theta_j$ on $q_j$. We make the following two observations.
% \begin{enumerate}
% \item Due to the normal form assumption, all the groundings of a first-order logic formula are identical to each other (up to renaming of a constant).  Hence, the resulting potential function
% $\Theta_j$'s are also identical to each other.
% \item If there are $r$ unary $\MAX$ predicates in $M_{\tilde{X}}$, then each $q_j$ can take $R=2^r$ possible values~\footnote{since there are $r$ predicate groundings for each $j$ and each is Boolean valued}.
% \end{enumerate}

\hspace{0.15cm}1)\;\;Due to the normal form assumption, all the groundings of a first-order logic formula behave identical to each other (up to renaming of constants).  Hence, the resulting potential function
$\Theta_j$'s are also identical to each other.

\hspace{0.15cm}2)\;\;If there are $r$ unary $\MAX$ predicates in $M_{\tilde{X}}$, then each $q_j$ can take $R=2^r$ possible values~\footnote{since there are $r$ predicate groundings for each $j$ and each is Boolean valued}.

%Using these observations, 
Therefore, the value of the product $\prod_{j=1}^{m} \Theta_j(q, s)$ in the RHS of Equation \ref{eqn:pt4} depends only on the number of different types of values $q_j$'s take in $q$ (and not on which $q_j$
takes which value). Let $\{v_1,v_2,\cdots,v_R\}$ denote the set of $R$ different values that $q_j$'s can take. Given a value $v_l$, let $N_l$ denote the number of times $v_l$ appears in $q$. Next, we state the following lemma.
\begin{lemma}\label{lem:pt4}
The MMAP problem can be written as:
\begin{align*}
\argmax_{q} W_{M_{\tilde X}}(q)=\argmax_{N_1,N_2,\cdots,N_R} \sum_{s} \prod_{l=1}^{R} f_l(s)^{N_l}
\end{align*}
subject to the constraints that $\forall l, N_l \ge 0, N_l \in \mathbb{Z}$ and $\sum_l N_l=m$. Here, $f_l(s)=\Theta_j(v_l,s)$.
\end{lemma}
\textit{Proof.} Proof follows from the fact that $\Theta_j$'s are symmetric to each other and that the $q_j$'s take a total of $m$ possible (non-unique) assignments since $\Delta_{\tilde{X}}=m$.

%	Since $\Theta_j$'s are identical to each other, $\prod_{j=1}^{m} \Theta'_j(q, s)$ only depends on the number of different kinds of assignments $q_j$'s take. Hence $\prod_{j=1}^{m} \Theta'_j(q, s)$
%	reduces to $\prod_{l=1}^{R}f_l(s)^{N_l}$.
We say that an assignment $N_1,N_2,\cdots,N_R$ subject to the constraints: $\forall l, N_l \ge 0$ and $\sum_l N_l=m$ is at {\em extreme} if $\exists l$ such that $N_l=m$. Note that for $R \ge 2$, extreme
assignment also implies that $\exists l, N_l=0$. We have the following lemma.
\begin{lemma}
\label{lem:pt4_eq_extreme}
	\hspace{-0.05in}{\bf *} The solution to the MMAP formulation $\argmax_{q} W_{M_{\tilde X}}(q)$ lies at extreme iff solution to its equivalent formulation:
	\begin{align*}
	\argmax_{N_1,N_2,\cdots,N_R} \sum_{s} \prod_{l=1}^{R} f_l(s)^{N_l}
	\end{align*}
	subject to the constraints $\forall l, N_l \ge 0, N_l \in \mathbb{Z}$ and $\sum_l N_l=m$ lies at extreme.
\end{lemma}
%The MMAP problem is now formulated as a maximization over discrete valued $N_l$ variables, reducing the complexity from exponential in $m$ to exponential in $L$, {\bf which can be significantly smaller}.
\vspace{-0.15cm}
\subsubsection{Proving Extreme}
\vspace{-0.15cm}
\begin{lemma}
\label{lem:convexity_extreme}
Consider the optimization problem:
	\begin{align*}
	\argmax_{N_1,N_2,\cdots,N_R} \sum_{s} g(s)\times \prod_{l=1}^{R} f_l(s)^{N_l}
	\end{align*}
	subject to the constraints $N_l \ge 0$, $\sum_l N_l=m$. $g(s)$ is an arbitrary real-valued function independent of $l$. The solution of this optimization problem lies at extreme.
\end{lemma}
% Convexity Proof start
% \begin{proof}
\textit{Proof.} Note that it suffices to prove this theorem assuming $N_l$'s are real-valued. If the solution is at extreme with real-valued $N_{l}$'s, it must also be at extreme when $N_l$'s are further constrained to be
integer valued. We will use induction on $R$ to prove the result. Consider base case of $R=2$, the function becomes $\argmax_{N_1} \sum_{s} {f_{1}(s)}^{N_1} {f_2(s)}^{m-N_1}\times g(s)$. This function is
convex and has its maximum value at $N_1=m$ or $N_1=0$ (see supplement for a proof).

Assuming that the induction hypothesis holds for $R=k$. We need to show for the case when $R=k+1$. We 
will prove it by contradiction. Assume that the solution to this problem does not lie at extreme. Then, in this solution, it must be the case that $N_l \neq 0, \forall l$. If not, we can then reduce the problem to a $k$ sized one and apply our induction hypothesis to get an extreme solution. Also, clearly $N_l < m, \forall l$. Let $N_{k+1}$ has the optimal value of $N^{*}_{k+1}$ in this solution. 
Then, substituting the optimal value of this component in the expression, we can get the optimal 
value for $(N_1,N_2,\cdots, N_k)$ by solving $\argmax_{N_1,N_2\cdots,N_k} \sum_{s} g'(s)\times \prod_{l=1}^{R} f_{l}(s)^{N_l}$, subject to $\sum_{l=1}^{k}N^l=m-N^{*}_{k+1}$. Here, $g'(s)=g(s)\times f_{k+1}(s)^{N^{*}_{k+1}}$. 
Using the induction hypothesis, the solution for this must be at extreme, i.e. $\exists l, N_l =0$ since $k \geq 2$. This is a contradiction.
% \end{proof}
\begin{corollary}\label{cor:pt4_extreme}
The solution to the optimization problem
	\begin{align*}
	\argmax_{N_1,N_2,\cdots,N_R} \sum_{s} \prod_{l=1}^{R} f_l(s)^{N_l}
	\end{align*}
	subject to the constraints $\forall l, N_l \ge 0$, $N_l \in \mathbb{Z}$ and $\sum_l N_l=m$  lies at extreme.
\end{corollary}
{\bf Theorem~\ref{thm:extreme} ({\it Proof}):} Corollary~\ref{cor:pt4_extreme} combined with Lemma~\ref{lem:pt4_eq_extreme}, Lemma~\ref{lem:pt4}, Lemma~\ref{lem:pt3}, Lemma~\ref{lem:pt2} and Lemma~\ref{lem:pt1} proves the theorem.

%Proof of this theorem follows from a combination of Lemma~\ref{lem:remove_unaryMAP}, Lemma \ref{lem:inversion_elim}, Lemma~\ref{lem:equiv_prodform}, Lemma \ref{lem:equal_extrema} and Lemma~\ref{lem:prod_extreme}.
\vspace{-0.25cm}
\subsection{SOM-R Rule for lifted MMAP}
\vspace{-0.25cm}
We will first define the SOM-R (SOM Reduce) equivalence class which is a sub-class of SOM. 
%In the following, we will assume that the MMAP problem is defined over an MLN $M$ with set of weighted
%formulas given as $\{f_i,w_i\}$. 
Following our notation, we will use $\mathcal{Q}$ and $\mathcal{S}$ to denote the set of \MAX and \SUM predicates, respectively in the MMAP problem.
%We will work with an MMAP problem defined over an MLN $M$
\begin{definition}
%Let $\mathcal{S}$ denote the set of \SUM predicates in the MMAP problem. 
We say that an equivalence class of variables $\tilde{X}$ is SOM-R if (a) $\tilde{X}$ is SOM (b) $\forall P \in
\mathcal{S}$, $P$ contains a variable from $\tilde{X}$ OR $\forall P \in \mathcal{S}$, $P$ does not have a variable from $\tilde{X}$.
\end{definition}
Note that if $|\mathcal{S}|=1$, then any SOM equivalence class is also necessarily SOM-R. Next, we exploit the properties of extreme assignments to show that domain of SOM-R variables can be reduced
to a single constant for MMAP inference. We start with the definition of a reduced MLN.
\begin{definition}\label{defn:reduceMLN}(Reduced MLN)
Let $\{(f_i,w_i\}_{i=1}^{n}$ denote the set of (weighted) formulas in $M$. Let $\tilde{X}$ be a SOM-R equivalence class with $|\Delta_{\tilde{X}}|=m$. We construct a reduced MLN $M^r$ by
considering the following 2 cases: \\ \\
CASE 1: $\forall P\in \mathcal{S}$,$P$ contains a variable from $\tilde{X}$
% \begin{itemize}
% \item $\forall f_i\in F(M)$ containing a variable $X\in\tilde{X}$, add $(w_i, f_{i})$ to $M^r$
% \item $\forall f_i\in F(M)$ not containing a variable $X\in\tilde{X}$, add $(\frac{1}{m}\times w_i,f_{i})$ to $M^r$. 
% \end{itemize}

\hspace{0.15cm}$\bullet$\;\;$\forall f_i\in F(M)$ containing a variable $X\in\tilde{X}$, add $\text{\hspace{0.5cm}}(f_{i}, w_i)$ to $M^r$.

\hspace{0.15cm}$\bullet$\;\;$\forall f_i\in F(M)$ not containing a variable $X\in\tilde{X}$, add $\text{\hspace{0.5cm}}(f_{i},\frac{1}{m}\times w_i)$ to $M^r$. \\

CASE 2: $\forall P\in \mathcal{S}$, $P$ does not contain a variable from $\tilde{X}$
% \begin{itemize}
% \item $\forall f_i\in F(M)$ containing a variable $X\in\tilde{X}$, add $(w_i\times m, f_{i})$ to $M^r$  
% \item $\forall f_i\in F(M)$ not containing a variable $X\in\tilde{X}$, add $(w_i,f_{i})$ to $M^r$. 
% \end{itemize}

\hspace{0.15cm}$\bullet$\;\;$\forall f_i\in F(M)$ containing a variable $X\in\tilde{X}$, add $\text{\hspace{0.5cm}}(f_{i},w_i\times m)$ to $M^r$.

\hspace{0.15cm}$\bullet$\;\;$\forall f_i\in F(M)$ not containing a variable $X\in\tilde{X}$, add $\text{\hspace{0.5cm}}(f_{i}, w_i)$ to $M^r$. 

%Case A: When the \SUM predicates contain a variable from $\tilde{X}$ (1)
%$\forall f_i \in F(M)$ containing a variable $X\in\tilde{X}$, add $(w_i*m, f_{i})$ to $M^r$ (2) $\forall f_i \in F(M)$ not containing a variable $X\in\tilde{X}$, add $(w_i,f_{i})$ to $M^r$. 
%Case B: When the \SUM predicates contain a variable from $\tilde{X}$ (1)
%$\forall f_i \in F(M)$ containing a variable $X\in\tilde{X}$, add $(w_i*m, f_{i})$ to $M^r$ (2) $\forall f_i \in F(M)$ not containing a variable $X\in\tilde{X}$, add $(w_i,f_{i})$ to $M^r$. 
%
In each case, we reduce the domain of $\tilde{X}$ to a single constant in $M^r$.
\end{definition}
%Revisiting MLN $M_1$, the reduced MLN $M_{1}^{R}$ with respect to the equivalence class $\tilde{Y}=\{Y,V\}$ (assuming $m=|\Delta_{\tilde{Y}}|$) is given as:
%\begin{align*}
%	&w_1 : \mathit{Frnds}(X,Y) \wedge \mathit{Parent}(Z,X) \Rightarrow \mathit{Knows}(Z,Y) \\
%	&w_2 : \mathit{Knows}(U,V) 
%\end{align*}
%where $w_1'=m*w_1$ and $w_2'=m*w_2$. The domain $\Delta_{\tilde{Y}}$ has been reduced to a single constant in $M_{1}^{R}$. 
%
We are ready to state our SOM-R rule for lifted MMAP.
%how that we can use $M^r$ to compute the MMAP assignment over the original MLN $M$.
%Our next lemma states that we can use the reduced MLN for MMAP computation when solutions lies at extreme for a given variable equivalence class.
\begin{theorem}\label{lem:reduce}({\em SOM-R Rule for MMAP})
Let $\tilde{X}$ be a SOM-R \eq class. Let $M^r$ be the reduced MLN in which domain of $\tilde{X}$ has been reduced to single constant. Then, MMAP problem can be equivalently solved over $M^r$.
% Let $\tilde{X}$ be a SOM-R \eq class. Let $M^r$ be the reduced MLN. Then, MMAP problem can be equivalently solved over $M^r$.
\end{theorem}
\textit{Proof.} Let $\mathcal{Q}$ denote the set of \MAX predicates in the problem. We prove the above theorem in two parts. In Lemma~\ref{lem:extreme_mapping} below, we show that for every extreme assignment (with respect to $\tilde{X}$)  $q$ to groundings of
$\mathcal{Q}$ in $M$, there is a corresponding extreme assignment $q^r$ in $M^r$ (and vice-versa).
%(lemma~\ref{lem:extreme_mapping}). 
In Lemma~\ref{lem:equal_mmap}, we show that given two extreme assignments, $q$ and $q^r$ for the respective MLNs, the MMAP value at $q$ (in $M_{\tilde{X}}$) is a monotonically increasing function of
the MMAP value at $q^r$ (in $M^r_{\tilde{X}}$). These two facts combined with the fact that MMAP solution to the original problem is at extreme (using Theorem~\ref{thm:extreme})
prove the desired result. Next we prove each result in turn.
\begin{lemma}\label{lem:extreme_mapping}
Let ${\bf q}$ (resp. ${\bf q^r}$) denote the sets of extreme assignments to the groundings of $\mathcal{Q}$ in $M$ (resp. $M^r$). There exists a one to one to mapping between ${\bf q}$ and ${\bf
q^r}$.
\end{lemma}
\textit{Proof.} Instead of directly working with $M$ and $M^r$, we will instead prove this lemma for the corresponding variablized MLNs $M_{\tilde{X}}$ and $M^{r}_{\tilde{X}}$. This can be done since the process
of variablization preserves the distribution as well as the set of extreme assignments. 
%Since the process of variablization
%preserves the distribution as well as set of extreme assignments (with respect to class $\tilde{X}$, showing a mapping of extreme assignments between $M_{\tilde{X}}$ and $M^r_{\tilde{X}}$ will prove the desired result.
%We note that $M^R_{\tilde{X}}$ is fully ground since $\Delta_{\tilde{X}}$ has been reduced to a single constant. 
Let $q$ denote an extreme assignment to \MAX predicates in $M_{\tilde{X}}$. We will construct a corresponding assignment $q^r$ for \MAX predicate in $M^r_{\tilde{X}}$. Since $\tilde{X}$ is SOM-R, $M_{\tilde{X}}$ has only unary and propositional predicates, whereas $M^r_{\tilde{X}}$ is full ground since the domain of
$\tilde{X}$  is reduced to a single constant. 

First, let us consider a propositional \MAX predicate $P$ in $M_{\tilde{X}}$. Since $P$ is ground both in $M$ and $M^r$, we can assign the value of $P$ in $q^r$ to be same as $q$. Next, let us consider a unary
predicate $P$. Let the assignments to the $m$ groundings of $P$ in $q$ be given by the set 
$\{q_{P_j}\}$ where $1 \le j \le m$. Since $q$ is extreme, each element in the set $\{q_{P_j}\}$ takes the same truth value. We can simply assign this value to the ground appearance of $P$ in
$M_{\tilde{X}}$. Hence, we get a mapping from $q$ to $q^r$. It is easy to see that we can get a reverse mapping from $q^r$ to $q$ in a similar manner. Hence, proved.\\
Next, we state the relationship between the MMAP values obtained by the extreme assignments in $M$ and $M^r$.
\begin{lemma}\hspace{-0.05in}{\bf *} 
\label{lem:equal_mmap}
Let $M$ be an MLN and $M^r$ be the reduced MLN with respect to the SOM-R equivalence class $\tilde{X}$. Let $q$ and $q^r$ denote two corresponding extreme assignments in $M$ and $M^r$, respectively. 
%e an extreme assignment to \MAX predicates groundings in $M$ and let $q^r$ denote the
%corresponding extreme assignment in $M^r$. 
Then, $\exists$ a monotonically increasing function $g$ such that $W_M(q)=g(W_{M^{r}}(q^r))$.
%where $W_M(q)$ and $W_{M^r}(q^r))$ are the MMAP values at $q$ (in $M$) and
%$q^r$ (in $M^r$), respectively. 
\end{lemma}
The proof of Lemma~\ref{lem:equal_mmap} exploits inversion elimination and symmetry of potential functions over a variablized MLN similar to their use in Section~\ref{sec:extreme}. These combined with Lemma~\ref{lem:extreme_mapping} become our key insights for reducing the complexity of MMAP inference significantly compared to existing methods (see supplement for details).
%Lemma~\ref{lem

%\begin{proof}
%({\it Intuition}) As earlier, we will work with the variablized MLNs $M_{\tilde{X}}$ and $M^r_{\tilde{X}}$. There are two kinds of formulas in $M_{\tilde{X}}$. First, for the formulas which do contain a
%variable from the set $\tilde{X}$, there is corresponding formula with the same weight in $M^r_{\tilde{X}}$. Both these formulas are ground. Second, for the formulas containing an occurrence of
%variables from the set ${\tilde{X}}$, there will be $m$ groundings in the network one for each constant. Further, $q$ being extreme, it will assign the same truth value to the \MAX predicates appearing in each of these groundings. In $M^r_{\tilde{X}}$, the number of these groundings is reduced to one but correspondingly, the weight of the formula has been multiplied by $m$ to mimic the $m$ (identical) groundings in $M^r_{\tilde{X}}$. Further, since $q$ and $q^r$ assign the same truth value to the
%corresponding \MAX predicates, we get equivalence between the two ground networks under the respective assignments. There is some added complication due to the presence of unary \SUM predicates which is
%handled using inversion elimination. We refer the reader to the supplement for a detailed proof.
%\end{proof}
\begin{corollary}
	SOM-R rule for MMAP problem subsumes SO rule for MAP problem given by Mittal et al.~\shortcite{mittal&al14}.
\end{corollary}
The corollary follows from the fact that MAP is a special case of MMAP when all the predicates are \texttt{MAX}. 
\vspace{-0.25cm}
\section{ALGORITHMIC FRAMEWORK}
\vspace{-0.25cm}
\label{sec:algo}
SOM-R rule can be combined with existing lifted inference rules such as lifted decomposition and conditioning \cite{jha&al10,gogate&domingos11} (with minor modifications) to yield a powerful algorithm for solving MMAP (see Algorithm~\ref{algo:lifted_mmap}).  The algorithm takes as input an MLN $M$, the set of \MAX predicates $\mathcal{Q}$, \SUM predicates $\mathcal{S}$ and a ground MMAP solver $\mathit{gSol}$.  It has six steps. In the first step, the algorithm checks to see if the MLN, along with $\mathcal{Q}$ and $\mathcal{S}$ can be partitioned into disjoint MLNs that do not share any ground atoms. If this condition is satisfied, then the MMAP solution can be constructed by solving each component independently and simply concatenating the individual solutions. In the next three steps, we apply the decomposer \cite{jha&al10}, SOM-R (this work) and binomial rules \cite{jha&al10,gogate&domingos11} in order. The former two reduce the domain of all logical variables in the equivalence class to a constant and thus yield exponential reductions in complexity. Therefore, they are applied before the binomial rule  which creates $O(m)$ ($|\Delta_{\tilde{X}}|=m$) smaller sub-problems. In the algorithm, $M^d$ refers to an MLN obtained from $M$ by setting the domain of $\tilde{X}$ to a single constant and we assume that $|\Delta_{\tilde{X}}|=m$. Similarly, $M^r$ refers to the MLN obtained from $M$ by applying the SOM-R rule (see Definition \ref{defn:reduceMLN}). 

The binomial rule (steps 4a and 4b) efficiently conditions on the unary predicates and can be applied over the \SUM as well as \MAX predicates. However, care must be taken to ensure that all \MAX predicates are instantiated before the \SUM predicates. Therefore, the binomial rule is applied over the \SUM predicates only when the MLN has no \MAX predicates (Step 4b). In the algorithm, $M_k$ refers to the MLN obtained from $M$ by setting exactly $k$ groundings of $P$ to true and the remaining to false. 

If none of the lifting rules are applicable and the MLN has only ground atom, we return the solution returned by the propositional solver $\mathit{gSol}$. Otherwise, if not all predicates are ground, we resort to partial grounding, namely we heuristically ground a logical variable and recurse on the corresponding MLN $M'$.

Finally, note that the algorithm returns the exponentiated weight of the MMAP assignment. The assignment can be recovered by tracing the recursion backwards.

%As the first step, we check for the existence of independent sub-theories. If so, the MMAP value can be computed independently for each one of them and then combined together. Next, we check for the application of lifting rules in the following order (a) decomposer~\cite{jha&al10,mittal&al15} (b) SOM (current work) (c) binomial~\cite{jha&al10,mittal&al15} on \MAX predicates. 

\eat{Previous literature~{} has observed ordering of the lifting rules plays an important rule in achieving good performance. In Algorithm~\ref{algo:lifted_mmap}, we first check for the applicability of
decomposer and SOM-R
since they reduce the domain to a single constant. On the other hand, the binomial rule creates $O(m)$ ($m=|\Delta_{\tilde{X}}|$) smaller sub-problems. Other rules such as domain
recursion~\cite{broeck&al11} could also be considered though in our experiments, it was never applicable (or superseded by the decomposer). If none of the lifting rules is applicable, we resort to
partial grounding and repeat the process since some lifting rules might now become applicable. 
}
{\bf Heuristics:} (a) Binomial: In case of multiple possible binomial applications, we pick the one which results in the application of other lifting rules (in the priority order described above) using a one
step look ahead. In case of a tie, we pick the one with maximum domain size.

(b) Partial Grounding: We pick the equivalence class which results in further application of lifting rules (in the priority order) using a one step look ahead. In case of a tie, we pick the one which has smallest domain size.
\vspace{-0.25cm}
\section{EXPERIMENTS}
\vspace{-0.25cm}
The goal of our experiments is two fold. First, we would like to examine the efficacy of lifting for MMAP. Second, we would like to analyze the contribution of SOM-R rule in lifting. Towards this end,
we compare the following three algorithms: (1) Ground: ground inference with no lifting whatsoever (2) Lifted-Basic: lifted inference without use of the SOM-R rule~\footnote{We use the rules described
in Algorithm~\ref{algo:lifted_mmap}. For Lifted-Basic, too many applications of the binomial rule led to blow up. So, we restricted the algorithm to a single binomial application and before any partial grounding. Lifted-SOM-R had no such issues.} 
(3) Lifted-SOM-R: using all our lifting rules including SOM-R. For ground inference, we use a publicly available~\footnote{https://github.com/radum2275/merlin} base (exact) solver built on top of And/Or search developed by Marinescu et al.~\shortcite{marinescu&al14}.
\setlength{\textfloatsep}{7pt}% Remove \textfloatsep
 \begin{algorithm}[t]
 \small
 \caption{Lifted MMAP}
 \textbf{Input:} MLN $M,\mathcal{Q},\mathcal{S},gSol$\\
 \textbf{Output:} MMAP value\\
 \label{algo:lifted_mmap}
 \begin{minipage}{\linewidth}
 \textbf{Begin:}
 \begin{algorithmic}
 \STATE //1. Disjoint Sub-Theories
 \IF {$M$ can be partitioned into disjoint MLNs $M_1,\ldots,M_t$ that share no atoms}
 	% \STATE return splitAndSolve($M$, $\mathcal{Q}$, $\mathcal{R}$, $gSol$))
 	\STATE {\bf return} $\prod_{i=1}^{t}$ liftedMMAP($M_i,\mathcal{Q}_i,\mathcal{S}_i$)
 \ENDIF
 \STATE //2. Decomposer
 \IF {there exists a decomposer $\tilde{X}$ in $M$}
 	\STATE {\bf return} [liftedMMAP($M^d,\mathcal{Q},\mathcal{S}$,gSol)]$^{m}$;
 \ENDIF
 \STATE //3. SOM-R (see Defn.~\ref{defn:reduceMLN})
 \IF {there exists a SOM-R class $\tilde{X}$ in $M$}
 	\STATE {\bf return} liftedMMAP($M^r,\mathcal{Q},\mathcal{S}$,gSol);
 \ENDIF
 \STATE //4a. Binomial over \MAX
 \IF {there exists a unary predicate $P\in \mathcal{Q}$}
	\STATE {\bf return} $\max_{k}$ liftedMMAP($M_k,\mathcal{Q}-\{P\},\mathcal{S}$,gSol);
 \ENDIF
 \STATE //4b. Binomial over \SUM
 \IF {$\mathcal{Q}=\emptyset$ and there exists a unary predicate  $P \in \mathcal{S}$} 
 	\STATE {\bf return} $\sum_{k=0}^{m}\;{m \choose k}$
 	 liftedMMAP($M_k,\mathcal{Q},\mathcal{S}-\{P\}$,gSol);
 \ENDIF 
 \STATE //5. Check if fully Ground
 \IF{$M$ is fully Ground}
   \STATE \textbf{return} apply($M',\mathcal{Q},\mathcal{S},gSol$);
  \ELSE
  \STATE //6. Partial Grounding
  \STATE $M'$ = Heuristically ground an equivalence class $\tilde{X}$ in $M$
   \STATE \textbf{return} liftedMMAP($M',\mathcal{Q},\mathcal{S},gSol$);
 \ENDIF
 \end{algorithmic}
 \textbf{End.}
 \end{minipage}
 \end{algorithm}

\begin{figure*}[!ht]
\small
\centering
\begin{minipage}[b]{0.3\textwidth}
\begin{subfigure}[]{1\textwidth}
	\includegraphics[width=\linewidth]{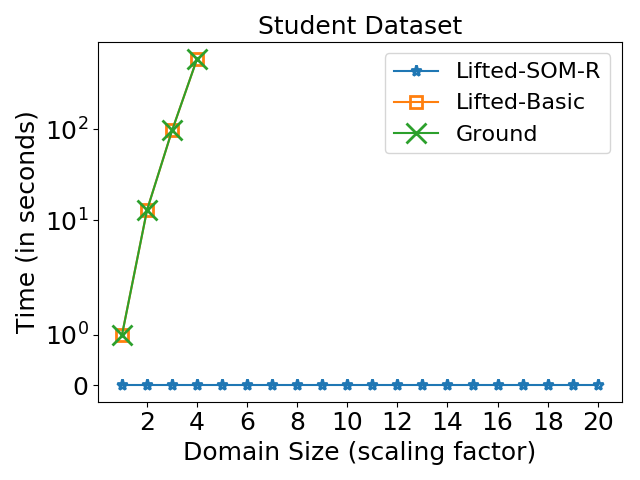}
	\caption{Student: time vs domain size}
	\label{fig:student_time}
\end{subfigure}
\end{minipage}
\begin{minipage}[b]{0.3\textwidth}
\begin{subfigure}[]{1\textwidth}
	\includegraphics[width=\linewidth]{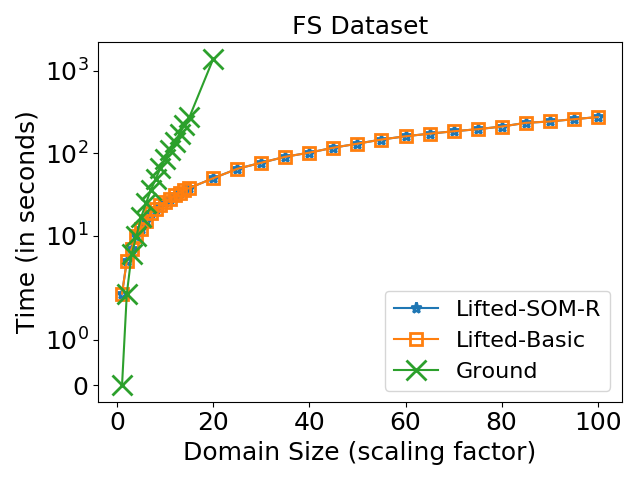}
	\caption{FS: time vs domain size}
	\label{fig:fs_time}
\end{subfigure}
\end{minipage}
\begin{minipage}[b]{0.3\textwidth}
\begin{subfigure}[]{1\textwidth}
	\includegraphics[width=\linewidth]{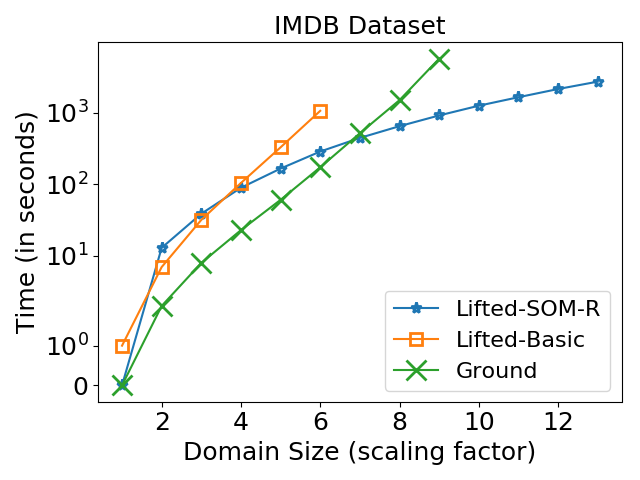}
	\caption{IMDB: time vs domain size}
	\label{fig:imdb_time}
\end{subfigure}
\end{minipage}
\begin{minipage}[b]{0.3\textwidth}
\begin{subfigure}[]{1\textwidth}
	\includegraphics[width=\linewidth]{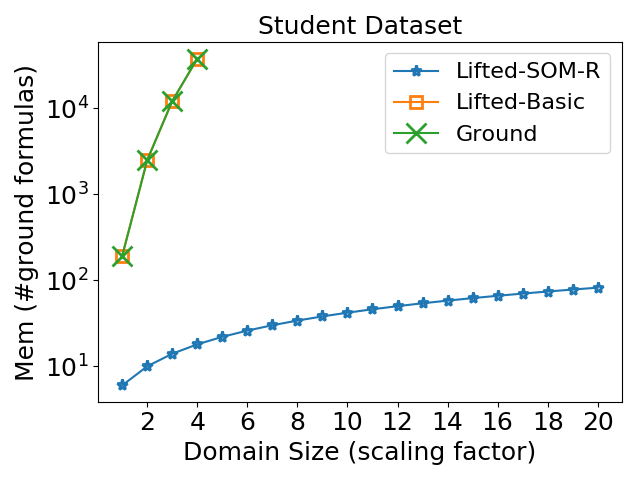}
	\caption{Student: mem vs domain size}
	\label{fig:student_mem}
\end{subfigure}
\end{minipage}
\begin{minipage}[b]{0.3\textwidth}
\begin{subfigure}[]{1\textwidth}
	\includegraphics[width=\linewidth]{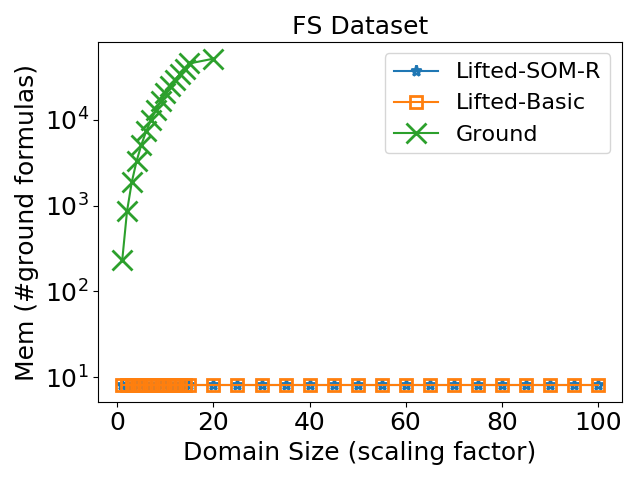}
	\caption{FS: mem vs domain size}
	\label{fig:fs_mem}
\end{subfigure}
\end{minipage}
\begin{minipage}[b]{0.3\textwidth}
\begin{subfigure}[]{1\textwidth}
	\includegraphics[width=\linewidth]{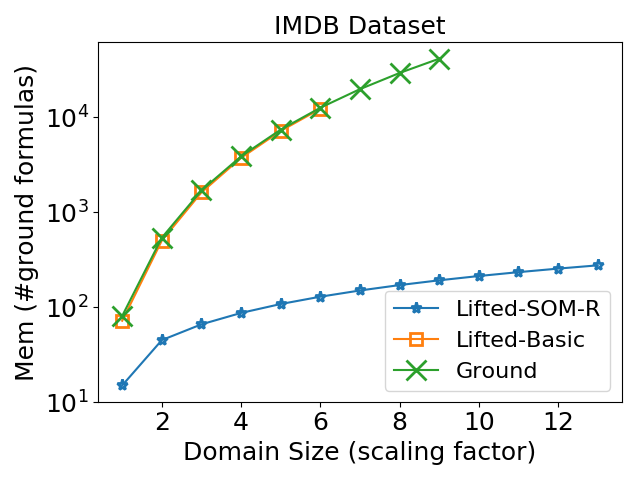}
	\caption{IMDB: mem vs domain size}
	\label{fig:imdb_mem}
\end{subfigure}
\end{minipage}
\begin{minipage}[b]{1\textwidth}
	\centering
	\begin{tabular}{l l}
		\begin{minipage}[b]{0.38\textwidth}
			\begin{tabular}{l l}
			\\\hline
			%\bf{Friends \& Smokers}~\cite{singla&domingos08}\\
			\bf{Student}~\cite{sarkhel&al14}\\
			\hline
			
			\specialcell[t]{
			Teaches(T, C) $\land$ Takes(S, C) $\Rightarrow$ JobOffer(S, M)} \\
			\textbf{MAP Predicate:} Takes(S, C), JobOffer(S, M)\\
			{\small \textbf{size:} teachr(T):2,course(C):3,comp(M):4,stud(S):6}\\

			\hline			
			\bf{FS}~\cite{domingos&lowd09}\\
			\hline
			\specialcell[t]{Smokes(P) $\Rightarrow$ Cancer({P});\\ 
			Smokes(P1) $\land$ Friend(P1, P2) $\Rightarrow$ Smokes(P2);}\\
			\textbf{MAP Predicates:} Smokes(P), Cancer(P)\\
			{\small \textbf{size:} person(P):5}\\
			\hline
			\end{tabular}
		\end{minipage}
		& 
		\begin{minipage}[b]{0.62\textwidth}
			\begin{tabular}{l l}
			\\\hline
			\bf{IMDB}~\cite{mittal&al16}\\
			\hline
			\specialcell[t]{
			WorksWith(P1,P2) $\Rightarrow$ Act(P1); WorksWith(P1,P2) $\Rightarrow$ Dir(P2);\\
			Dir(P1) $\land$ Act(P2) $\land$ Mov(M,P1) $\land$  Mov(M,P2) $\Rightarrow$ WorksWith(P2,P1);\\
			Dir(P1) $\land$ Act(P2) $\land$ Mov(M,P2) $\land$ WorksWith(P2,P1) $\Rightarrow$ Mov(M,P1);\\
			Dir(P1) $\land$ Act(P2) $\land$ Mov(M,P1) $\land$ WorksWith(P2,P1) $\Rightarrow$ Mov(M,P2);\\
			Dir(P1) $\land$ Act(P2) $\Rightarrow$ WorksWith(P2,P1);\\
			}\\
			\textbf{MAP Predicates:} Act(P), Dir(P), Mov(M,P)\\
			{\small \textbf{size:} person(P):3, movie(M):2}\\\\
			\hline
			\end{tabular}
		\end{minipage}
	\end{tabular}
\end{minipage}

\caption{Results and rules of Student, FS and IMDB datasets. "size" gives initial domain sizes for each case.}
% \caption{Results and rules of Student, IMDB and F\&S datasets. "Dom" in last row is the initial domain size for variables.}
\label{fig:Results_Rules}
\end{figure*}

We experiment with three benchmark MLNs: (1) Student~\cite{sarkhel&al14} (2) IMDB~\cite{mittal&al16} (3) Friends \& Smokers (FS)~\cite{domingos&lowd09}. All the datasets are described
in the lower part of Figure~\ref{fig:Results_Rules} along with the MAP predicates used in each case; the remaining predicates are treated as marginal predicates. Weights of the formulas were manually set. 

We compare the performance of the three algorithms on two different metrics: (a) time taken for inference (b) memory used. We used a time-out of 30 minutes for each run. Memory was measured in terms of the number of formulas in the ground network in each case. We do not compare the solution quality since all the algorithms are guaranteed to produce MMAP assignments with same (optimal) probability. All the experiments were run on a 2.20 GHz Xeon(R) E5-2660 v2 server with 10 cores and 62 GB RAM.

%In each case, we compare the performance of the three algorithms as we vary the domain size. x-axis plots the domain size and y-axis plots time (memory) in each case. y-axis is plotted on log-scale.

{\bf Results:} For each of the graphs in Figure~\ref{fig:Results_Rules}, we plot time (memory) on y-axis (log-scale) and domain size on x-axis. Time is measured in seconds. Since we are primarily concerned about
the scaling behavior, we use the number of ground formulae as a proxy for the actual memory usage. Domain size is measured as a function of a scaling factor, which is the number by which (all of) the
starting domain sizes are multiplied. We refer to domain descriptions (Figure~\ref{fig:Results_Rules}) for the starting sizes. %y-axis is plotted on log scale in all the graphs.

Figures~\ref{fig:student_time} and ~\ref{fig:student_mem} compare the performance of the three algorithms on the Student dataset. None of the lifting rules apply for Lifted-Basic. Hence, its
performance is identical to Ground. For Lifted-SOM-R, all the variables (except teacher(T)) can be reduced to a single constant, resulting in significant reduction in the size of the ground theory. 
%in constant time (and memory) performance. 
Lifted-SOM-R is orders of magnitude better than Ground and Lifted-Basic for both time and memory.
%can b
%We start with initial domain size $2, 3, 4, 6$ for $teacher$ ($t$), $course$ ($c$), $company$ ($m$) and $student$ ($s$) variables, respectively. 

Figures~\ref{fig:fs_time} and ~\ref{fig:fs_mem} compare the three algorithms on the FS dataset. Here, Lifted-Basic performs identical to Lifted-SOM-R. This is because binomial rule applies in the
beginning on Smokes, following which theory decomposes. We never need to apply SOM-R rule on this domain. Both Lifted-SOM-R and Lifted-Basic perform significantly better than Ground on this domain (in
both time and memory).

IMDB dataset (Figures~\ref{fig:imdb_time} and ~\ref{fig:imdb_mem}) presents a particularly interesting case of interspersed application of rules. For Lifted-SOM-R, SOM-R rule applies on movie(M)
variables, simplifying the theory following which binomial rule can be applied on Mov, Dir and Act predicates. Theory decomposes after these binomial applications. For Lifted-Basic, though binomial
rule can be applied on Dir, Act the movie variables still remain, eventually requiring for partial grounding. Surprisingly, Ground does slightly better than both the lifted approaches for smaller
domains for time. This is due to the overhead of solving multiple sub-problems in binomial without much gain since domains are quite small. Lifted-SOM-R has a much better scaling behavior
for larger domains. It also needs significantly less memory compared to both other approaches.

In none of the above cases, Lifted-SOM-R has to ever partially ground the theory making a very strong case for using Lifted-SOM-R for MMAP inference in many practical applications. Overall, our experiments clearly demonstrate the utility of SOM-R in the scenarios where other lifting rules fail to scale.

\vspace{-0.25cm}
\section{CONCLUSION}
\vspace{-0.25cm}
We present the first lifting technique for MMAP. Our main contribution is the SOM-R rule, which states that the domain of a class of equivalence variables, referred to
as SOM-R, can be reduced to a single constant for the purpose of MMAP inference. We prove the correctness of our rule through a series of problem transformations followed by the properties of what we refer to as extreme assignments. Our experiments clearly demonstrate the efficacy of our approach on benchmark domains.
Directions for future work include coming up with additional lifting rules, approximate lifting and lifting in presence of constraints~\cite{mittal&al15}, all in the context of MMAP, and experimenting with a wider set of domains.
\vspace{-0.5cm}
\subsubsection*{Acknowledgements}
\vspace{-0.15cm}
Happy Mittal is supported by the TCS Research Scholars Program.
Vibhav Gogate and Parag Singla are supported by the DARPA Explainable Artificial Intelligence (XAI) Program with number N66001-17-2-4032. Parag Singla is supported by IBM Shared University Research Award and the Visvesvaraya Young Faculty Research Fellowship by the Govt. of India.
Vibhav Gogate is supported by the National Science Foundation grants IIS-1652835 and IIS-1528037.
Any opinions, findings, conclusions or recommendations expressed in this paper are those of the authors and do not necessarily reflect the views or official policies, either expressed or implied, of the funding agencies.
% 
% 
% Happy Mittal is being supported by the TCS Research Scholars Program.
% Vibhav Gogate and Parag Singla are being supported by the DARPA Explainable Artificial Intelligence (XAI) Program with number N66001-17-2-4032. Parag Singla is being supported by IBM Shared University Research Award and the Visvesvaraya Young Faculty Research Fellowship by the Govt. of India.
% Vibhav Gogate is being supported by the National Science Foundation grants IIS-1652835 and IIS-1528037.
% Any opinions, findings, conclusions or recommendations expressed in this paper are those of the authors and do not necessarily reflect the views or official policies, either expressed or implied, of the funding agencies.
\clearpage
\newpage
\bibliographystyle{uai2018}
\bibliography{all}
\clearpage
\newpage
\section*{Supplementary Material}
\section*{Lemmas}
\label{sup_sec_lemmas}
We will start by proving Lemma \ref{lem:fxy}, which will be used in the proof of Lemma \ref{lem:remove_unaryMAP}
\setcounter{lemma}{-1}.
\begin{lemma}
    \label{lem:fxy} 
    Given a function $f(x,y)$, let $(x^*,y^*)$ be a maximizing assignment, i.e., $(x^{*},y^{*})= 
    \argmax_{x,y}f(x,y)$. Then, $\forall y'$ s.t. $y^{'} = \argmax_{y}f(x^*,y)$, $(x^*,y^{'})$ is also 
    a maximizing assignment.
    \end{lemma}
    \textit{Proof.} We can write the following (in)equality:
        \begin{align}
            % f(x^*,y^*) &\leq \underset{y}{\argmax}f(x^*,y) = f(x^*,y^{\prime *})
            f(x^*,y^*) &\leq \underset{y}{\max} f(x^*,y) = f(x^*,y^{\prime})\nonumber
        \end{align}
    But since $(x^{*}$,$y^{*})$ was the maximizing assignment for $f(x,y)$, it must be the case that $f(x^*,y^*)=f(x^*,y')$. Hence, $(x^{*},y')$ must also be a maximizing assignment. Hence, proved.

\setcounter{lemma}{1}
\begin{lemma}
\label{lem:remove_unaryMAP}
\hspace{-0.05in} Consider the MMAP problem over $M_{\tilde{X}}$. Let $q_{p}$ be an assignment 
to the propositional MAP predicates. Let $M^{\prime}_{ \tilde{X}}$ be an MLN obtained by substituting the 
truth value in $q_{p}$ for propositional predicates. Then, if $M^{\prime}_{ \tilde{X}}$ has a solution at 
extreme for all possible assignments of the form $q_{p}$ then, $M_{\tilde{X}}$ also has a solution at 
extreme.
\end{lemma}
\textit{Proof.} The MMAP problem can be written as:    
  \begin{align}
        \argmax_{q_p,q_{u}} \sum_{s_{p},s_{u}} W_{M_{\tilde{X}}}{(q_p,q_{u},s_{p},s_{u})}
    \end{align}
here, $q_p,q_u$ denote an assignment to the propositional and unary \MAX predicate groundings in $M_{\tilde{X}}$, respectively. Similarly, $s_p,s_u$ denote an assignment to the propositional and unary \SUM predicate
groundings in $M_{\tilde{X}}$, respectively. Let $q^{*}_{p}$ denote an optimal assignment to the propositional \MAX predicates. Then, using Lemma~\ref{lem:fxy}, we can get the MMAP assignment $q^{*}_u$ as a solution to the following problem:
%Proof about removing propositional MAP predicates
    \begin{align}
        \argmax_{q_{u}} \sum_{s_{p},s_{u}} W_{M_{\tilde{X}}^{\prime}}{(q_{u},s_{u},s_{p})}
    \end{align}
    where $M'_{\tilde{X}}$ is obtained by substituting the truth assignment $q^{*}_{p}$ in $M_{\tilde{X}}$. Since, $M'_{\tilde{X}}$ has a solution at extreme $\forall q_p$, it must also be at extreme
  when $q_p=q^{*}_p$. Hence, $q^{*}_u$ must be at extreme. Hence, proved.

\setcounter{lemma}{4}
\begin{lemma}
\label{lem:equal_extrema}
    % \hspace{-0.05in}$\argmax_{q_{u}} W_{M'_{\tilde X}}(q_{u})$ lies at extreme iff $\argmax_{N_1,N_2,\cdots,N_{\mathcal{L}}} \sum_{s_{p}} \prod_{l=1}^{\mathcal{L}} \Gamma_l(s_{p})^{N_l}$ s.t. $\sum_l N_l=m$ lies at extreme i.e.\ $\exists$ a MAP solution $N^{*}=(N_1^{*},N_2^{*},\cdots,N^{*}_{\mathcal{L}})$ such that $\exists l$, $N^{*}_l = m$ and $N^{*}_{l^{\prime}}=0$, $\forall {l^{\prime}} \neq l$.
    The solution to the MMAP formulation $\argmax_{q} W_{M_{\tilde X}}(q)$ lies at extreme iff solution to its equivalent formulation:
  \begin{equation}
  \argmax_{N_1,N_2,\cdots,N_R} \sum_{s} \prod_{l=1}^{R} f_l(s)^{N_l}
  \end{equation}
  subject to the constraints $\forall l, N_l \ge 0, N_l \in \mathbb{Z}$ and $\sum_l N_l=m$ lies at extreme.
\end{lemma}
\textit{Proof.} If $\argmax_{N_1,N_2,\cdots,N_R} \sum_{s} \prod_{l=1}^{R} f_l(s)^{N_l}$ lies at extreme then $\exists l$ such that $N_l=m$ and $N_{l'}=0, \forall l' \neq l$. Let $v_l$ denote the value taken by the
   groundings of the unary \MAX predicates corresponding to index $l$. Since $N_l=m$, it must be the case that all the groundings get the identical value $v_l$. Hence, the solution to $\argmax_{q}
   W_{M_{\tilde X}}(q)$ lies at extreme. Similar proof strategy holds for the other way around as well.
%Proof of Convexity. 
% \renewcommand\thetheorem{\Alph{theorem}}
% \setcounter{lemma}{}
\begin{lemma*}
\label{convexity}
    {\bf $\textit{(Induction Base Case in Proof of Lemma 6)}$}:  Let $f_{1}(s), f_{2}(s)$ and $g(s)$ denote real-valued functions of a vector valued input $s$~\footnote{Recall that $s$ was an assignment to all
  the propositional \SUM predicates in our original Lemma.}. Further, let each of $f_1(s),f_2(s),g(s)$ be non-negative. Then, for $N \in \mathcal{R}$, we define a function $h(N) = \sum_{s} {f_{1}(s)}^{N}
  {f_2(s)}^{m-N}\times g(s)$ where the domain of $h$ is further restricted to be in the interval $[0,m]$, i.e., $0 \le N \le m$. The maxima of the function $h(N)$ lies at  $N=0$ or $N=m$.
\end{lemma*}
\textit{Proof.} First derivative of $h(N)$ with respect to $N$ is:
\begin{align*}
    \frac{dh}{d N} = \sum_{s}\Big( &{f_{1}(s)}^{N}{f_{2}(s)}^{m-N}{g(s)}\times\\
    &{[log(f_{1}(s)) - log(f_{2}(s))]}\Big)
\end{align*}
Second derivative of $h(N)$ with respect to $N$ is given as:
\begin{align*}
    \frac{d^2h}{d N^2} = \sum_{s}\Big( &{f_{1}(s)}^{N}{f_{2}(s)}^{m-N}{g(s)}\times\\   
    &{[log(f_{1}(s)) - log(f_{2}(s))]}^2\Big)\geq 0\\
\end{align*}
The inequality follows from the fact that each of $f_1,f_2,g$ is non-negative. Hence, the second derivative of $h(N)$ is non-negative which means the function is convex. Therefore, the maximum value of this function must lie at the end points of its domain, i.e, either at $N=0$ or at $N=m$.

% ==============================================================================
% ==============================================================================
% ==============================================================================
% ==============================================================================
% ==============================================================================
% ==============================================================================
% ==============================================================================
% ==============================================================================
 
%\setcounter{lemma}{100}
%\begin{lemma}
%\label{lemma:weight_mul}
%  Let $M$ be an MLN and $(w,f)\in F(M)$ be a formula in $M$. Let $(w^\prime, f)$ be a new formula such that $w^\prime=w*m$ where $m$ is a constant. Let $\phi$ and $\phi^\prime$ be the potential functions over $f$ and $f^\prime$. Then we can write $\phi^\prime$ as,
%  \begin{align*}
%    \phi^\prime=\phi^m
%  \end{align*}
%\end{lemma}
%\begin{proof}
%  For any assignment $t$, $\phi(t)= exp^{wf(t)}$. Next, for the same assignment $t$,
%  \begin{align*}
%   \phi^\prime(t)&=exp^{m*wf(t)}\\
%   \phi^\prime(t)&=(exp^{wf(t)})^m\\
%   \phi^\prime(t)&=(\phi(t))^m
%  \end{align*}
%\end{proof}
\setcounter{lemma}{7}
% \begin{theorem}
% Given an MLN $M$ and single occurrence class $\tilde{X}$, let $M^R$ be the reduced MLN corresponding 
% to $\tilde{X}$. Then, MMAP problem over $M$ can be reduced to MMAP problem over $M^R$.
% \end{theorem}
\begin{lemma}
\label{lem:equal_mmap}
Let $M$ be an MLN and $M^r$ be the reduced MLN with respect to the SOM-R equivalence class $\tilde{X}$. Let $q$ and $q^r$ denote two corresponding extreme assignments in $M$ and $M^r$, respectively. 
Then, $\exists$ a monotonically increasing function $g$ such that $W_M(q)=g(W_{M^{r}}(q^r))$.
\end{lemma}
\textit{Proof.} First, we note that if we multiply the weight $w_i$ of a formula $f_i$ in
an MLN by a factor $k$, then, the corresponding potential $\phi_{ij}$ (i.e., potential corresponding to the $j^{th}$ grounding of the $i^{th}$ formula) gets raised to the power $k$. If $w_i$ gets
replaced by $k \times w_i$, then, correspondingly, $\phi_{ij}$ gets replaced by $(\phi_{ij})^k$ $\forall j$. We will use this fact in the following proof.

As in the case of Lemma 7, we will instead work with the variablized MLNs $M_{\tilde{X}}$ and $M^{r}_{\tilde{X}}$, respectively.  
%In this proof we will work with variablized forms of $M$ and $M^r$. $M_{\tilde{X}}$ and $M^r_{\tilde{X}}$ rather than $M$ and $M^r$ as an MLN and its variablized form represent same distribution.
Let $q=(q_p, q_u)$ be the MMAP assignment for $\mathcal{Q}$ in $M_{\tilde{X}}$ and similarly $q=(q^r_p, q^r_u)$ be the MMAP assignment for $\mathcal{Q}$ in $M^r_{\tilde{X}}$.

For MLN $M_{\tilde{X}}$, the MMAP objective $W_M$ at $(q_p, q_u)$ can be written as $W_{M_{\tilde{X}}}(q_{p}, q_{u})=:$
\begin{align}
 \sum_{s_{p}, s_{u}}\Big( \prod_{i=1}^{r} \prod_{j=1}^{m} \phi_{ij}(q_{p}, q_{u}, s_{p}, s_{u}) \prod_{k=1}^{t} \phi_k(q_{p}, s_{p})\Big)\label{equation:potential}\\
 \nonumber
\end{align}
where $\phi_{ij}$ are potentials over formulas containing some $X\in\tilde{X}$ and $\phi_{k}$ are potentials over formulas which do not contain any $X\in\tilde{X}$. In particular, note that we have
separated out the formulas which involve a variable from the class $\tilde{X}$ from those which don't. 
$r$ denotes the count of the formulas of the first type and $t$ denotes the count of the formulas of the second type. We will use this form in the following proof.

Let the reduced domain of $\tilde{X}$ in $M^r$ is given by $\{x_1\}$, i.e., the only constant which remains in the domain is corresponding to index $j=1$. Next we prove the above lemma for the two cases considered in Definition 5:\\

{\bf CASE 1}: $\forall P\in \mathcal{S}$,$P$ contains a variable from $\tilde{X}$\\
In this case $M_{\tilde{X}}$ and $M^r_{\tilde{X}}$ will not contain any propositional \SUM predicate i.e. $s_p=\emptyset$.

In this case, while constructing $M^r_{\tilde{X}}$ , for formulas not containing some $X\in \tilde{X}$ we divided the weight by $m$. This combined with the result stated in the beginning of this
proof, the MMAP objective for $M^r_{\tilde{X}}$ can be written as:
\begin{align*}
   W_{M^r_{\tilde{X}}}(q_{p}, q_{u})=&\sum_{s_{u}}\Big( \prod_{i=1}^{r} \phi_{i1}(q_{p}, q_{u_1}, s_{u_1}) \prod_{k=1}^{t} \phi_k(q_{p})^{\frac{1}{m}}\Big)\nonumber\\
   =&\Big(\sum_{s_{u_1}} \prod_{i=1}^{r} \phi_{i1}(q_{p}, q_{u_1}, s_{u_1})\Big) \prod_{k=1}^{t} \phi_k(q_{p})^{\frac{1}{m}}
 \end{align*}

Next for MLN $M_{\tilde{X}}$ we have, $W_{M_{\tilde{X}}}(q_{p}, q_{u})=$
\begin{align}
    &\sum_{s_{u}}\Big( \prod_{i=1}^{r} \prod_{j=1}^{m} \phi_{ij}(q_{p}, q_{u}, s_{u}) \prod_{k=1}^{t} \phi_k(q_{p})\Big)\nonumber\\
    =&\Big(\sum_{s_{u}} \prod_{j=1}^{m} \prod_{i=1}^{r} \phi_{ij}(q_{p}, q_{u}, s_{u}) \Big) \prod_{k=1}^{t} \phi_k(q_{p})\nonumber
\end{align}
\begin{align}    
    =&\Big(\sum_{s_{u_j}} \prod_{j=1}^{m} \prod_{i=1}^{r} \phi_{ij}(q_{p}, q_{u_j}, s_{u_j}) \Big)\prod_{k=1}^{t} \phi_k(q_{p})\nonumber\\
    =&\Big(\prod_{j=1}^{m} \sum_{s_{u_j}} \prod_{i=1}^{r} \phi_{ij}(q_{p}, q_{u_j}, s_{u_j}) \Big)\prod_{k=1}^{t} \phi_k(q_{p})\nonumber\\
    =&\Big(\prod_{j=1}^{m} \sum_{s_{u_j}} \prod_{i=1}^{r} \phi_{ij}(q_{p}, q_{u_1}, s_{u_j}) \Big)\prod_{k=1}^{t} \phi_k(q_{p})\nonumber\\
    =&\Big(\sum_{s_{u_j}} \prod_{i=1}^{r} \phi_{ij}(q_{p}, q_{u_1}, s_{u_j}) \Big)^m\prod_{k=1}^{t} \phi_k(q_{p})\nonumber\\
    =&\Big(\sum_{s_{u_j}} \prod_{i=1}^{r} \phi_{ij}(q_{p}, q_{u_1}, s_{u_j}) \Big)^m\prod_{k=1}^{t} \Big(\phi_k(q_{p})^{\frac{1}{m}}\Big)^m\nonumber\\
    =&\Big(\sum_{s_{u_j}} \prod_{i=1}^{r} \phi_{ij}(q_{p}, q_{u_1}, s_{u_j}) \Big)^m \Big(\prod_{k=1}^{t} \phi_k(q_{p})^{\frac{1}{m}}\Big)^m\nonumber\\
    =&\Big(\sum_{s_{u_j}} \prod_{i=1}^{r} \phi_{ij}(q_{p}, q_{u_1}, s_{u_j})\prod_{k=1}^{t} \phi_k(q_{p})^{\frac{1}{m}}\Big)^m\nonumber\\
    =&\Big(W_{M^r_{\tilde{X}}}(q_{p}, q_{u})\Big)^m\nonumber
\end{align}

First equality comes by removing $s_p$ from Equation \ref{equation:potential}. In second equality we switch the order of two products. In third equality we have made explicit the dependence of
$\phi_{ij}$ on $q_{u_j}$ and $s_{u_j}$ i.e. groundings corresponding to $j^{th}$ constant. In fourth equality we use inversion elimination~\cite{braz&al05} to invert the sum over $s_{u_j}$ and product
over $j$. Next, since $\tilde{X}$ is SOM-R, from Theorem 1 we know $q_u$ lies at extreme i.e. $\forall j$, $q_{u_j} = q_{u_1}$, so we replace all $q_{u_j}$ by $q_{u_1}$ in fifth equality. Next, after summing out $s_{u_j}$ all
$\phi_{ij}$ will behave identically~\footnote{$\phi_{ij}$'s are identical to each other up to renaming of the index $j$, due to the normal form assumption.}, so we reduce $\prod_{j=1}^{m}$ to exponent $m$. In the next
steps we do basic algebraic manipulations to show $W_{M_{\tilde{X}}}(q)=g(W_{M^{r}_{\tilde{X}}}(q^r))$ where $g$ is
function defined as $g(x)=x^m$, i.e., $g$ is monotonically increasing. Hence, proved.

{\bf CASE 2}: $\forall P\in \mathcal{S}$, $P$ doesn't contain a variable from $\tilde{X}$\\
In this case $M_{\tilde{X}}$ will not contain any unary \SUM predicate i.e. $s_u=\emptyset$.

In this case for the reduced MLN $M^r_{\tilde{X}}$ we multiply the weight of formulas containing some $X\in \tilde{X}$ by $m$ and domain of $\tilde{X}$ is reduced to a single constant. Combining this
fact along with the result shown in the beginning of this proof, MAP objective for $M^r_{\tilde{X}}$ is given by:
\begin{align*}
   W_{M^r_{\tilde{X}}}(q_{p}, q_{u})=&\sum_{s_{p}}\Big( \prod_{i=1}^{r} \phi_{i1}(q_{p}, q_{u_1}, s_{p})^m \prod_{k=1}^{t} \phi_k(q_{p}, s_{p})\Big)
   % &=W_{M_{\tilde{X}}}(q_{p}, q_{u}) \quad\quad\quad(\text{from Equation \ref{equation:final_case2}})
 \end{align*} 

Next for MLN $M_{\tilde{X}}$ we have, $W_{M_{\tilde{X}}}(q_{p}, q_{u})=$
\begin{align}
    &\sum_{s_{p}}\Big( \prod_{i=1}^{r} \prod_{j=1}^{m} \phi_{ij}(q_{p}, q_{u}, s_{p}) \prod_{k=1}^{t} \phi_k(q_{p}, s_{p})\Big)\nonumber\\
    &=\sum_{s_{p}}\Big( \prod_{i=1}^{r} \prod_{j=1}^{m} \phi_{ij}(q_{p}, q_{u_j}, s_{p}) \prod_{k=1}^{t} \phi_k(q_{p}, s_{p})\Big)\nonumber\\
    &=\sum_{s_{p}}\Big( \prod_{i=1}^{r} \prod_{j=1}^{m} \phi_{ij}(q_{p}, q_{u_1}, s_{p}) \prod_{k=1}^{t} \phi_k(q_{p}, s_{p})\Big)\nonumber\\
    &=\sum_{s_{p}}\Big( \prod_{i=1}^{r} (\phi_{i1}(q_{p}, q_{u_1}, s_{p}))^m \prod_{k=1}^{t} \phi_k(q_{p}, s_{p})\Big)\nonumber\\
    &=W_{M^r_{\tilde{X}}}(q_{p}, q_{u})\nonumber
\end{align}
First equality comes by removing $s_u$ from Equation \ref{equation:potential}. In second equality we have made explicit the dependence of $\phi_{ij}$ on $q_{u_j}$ i.e. groundings corresponding to
$j^{th}$ constant. Next, since $\tilde{X}$ is SOM-R, from Theorem 1 we know $q_u$ lies at extreme i.e. $\forall j$, $q_{u_j} = q_{u_1}$, so we replace all $q_{u_j}$ by $q_{u_1}$ in third equality. Last equality comes from the
fact that $\phi_{ij}$'s are identical to each other up to renaming of the index $j$ as argued earlier.
%, due the normal form assumption.
%fact that given $q_{p}, q_{u_1} \text{ and } s_{p}$, $\phi_{ij}$ behave identical and independent of $j$, 
Hence we can write $\prod_j \phi_{ij}$ as $(\phi_{i1})^m$. Hence, in this case, we have
$W_{M_{\tilde{X}}}(q)=W_{M^r_{\tilde{{X}}}}(q^r)$ implying that the function $g$ is identity (and hence, monotonically increasing).

From proofs of Case 1 and Case 2 we conclude that $\exists$ a monotonically increasing function $g$ such that $W_M(q)=g(W_{M^{r}}(q^r))$.
\end{document}